\documentclass{arxiv}

\usepackage{times}
\usepackage{booktabs}
\usepackage{hyperref}
\usepackage{amsmath}
\usepackage{amssymb}
\usepackage{caption}

\numberwithin{figure}{section}
\numberwithin{table}{section}

\title[Ontology-Guided Neuro-Symbolic Inference]{Ontology-Guided Neuro-Symbolic Inference: Grounding Language Models with Mathematical Domain Knowledge}

\author{%
 \Name{Marcelo Labre} \Email{marcelo.labre@advancedinstitute.ai}\\
 \addr Advanced Institute for Artificial Intelligence (AI2)%
}

\begin{document}

\makeatletter
\def\@setjmlr{\vspace*{-32pt}}
\def\jmlr@manuscript{}
\makeatother

\raggedbottom  

\maketitle

\begin{abstract}
Language models exhibit fundamental limitations---hallucination, brittleness, and lack of formal grounding---that are particularly problematic in high-stakes specialist fields requiring verifiable reasoning. I investigate whether formal domain ontologies can enhance language model reliability through retrieval-augmented generation. Using mathematics as proof of concept, I implement a neuro-symbolic pipeline leveraging the OpenMath ontology with hybrid retrieval and cross-encoder reranking to inject relevant definitions into model prompts. Evaluation on the MATH benchmark with three open-source models reveals that ontology-guided context improves performance when retrieval quality is high, but irrelevant context actively degrades it---highlighting both the promise and challenges of neuro-symbolic approaches.
\end{abstract}

\begin{keywords}
Neuro-symbolic AI, Ontology-guided Inference, Language Models, Knowledge-enriched Learning, Retrieval-Augmented Generation, Domain Grounding
\end{keywords}

\section{Introduction}
\label{sec:introduction}

The deployment of language models in specialist domains---medicine, finance, law, and scientific research---has revealed a fundamental tension between impressive pattern-matching capabilities and the rigorous requirements of high-stakes decision-making. Language models ``interpret'' domain tasks through learned statistical patterns rather than ``compiling'' them against ground truth, explaining why their outputs cannot be trusted where accuracy and traceability are paramount.

Consider the contrast with software development, where code generation success stems from external verification mechanisms: compilers, type systems, and unit tests provide ground truth against which generated code can be validated. The question motivating this research is whether similar verification mechanisms can be created for other specialist domains.

I argue that formal ontologies represent a promising foundation for such infrastructure. An ontology provides structured domain knowledge---concepts, relationships, properties, and axioms---that can serve as ground truth for validating language model outputs. Unlike implicit knowledge embedded in model parameters, ontological knowledge is explicit, machine-readable, and maintainable by domain experts. This research deliberately focuses on small to medium language models (9B parameters and smaller), aiming to demonstrate that even smaller models can achieve meaningful improvements when guided by formal ontological knowledge.

I investigate this vision using mathematics as a proof-of-concept domain, exploring the OpenMath standard---a W3C-aligned specification for representing mathematical semantics through Content Dictionaries that define symbols with machine-readable Formal Mathematical Properties. The core hypothesis is that integrating formal mathematical ontologies with language model inference can improve reasoning reliability when retrieval successfully identifies the ontology's relevant definitions---and conversely, that irrelevant symbols create noise that degrades performance.

\section{Background}

\subsection{Literature Review}

The integration of neural and symbolic approaches has emerged as a significant research direction across high-stakes domains \citep{kautz2022third,sarker2021neuro}. In healthcare, systems like RAAPID integrate Clinical Knowledge Graphs with language models for risk assessment \citep{raapid2026}, while legal applications such as NeuReg combine regulatory knowledge structures with neural generation to reduce hallucination in compliance-critical outputs \citep{arshad2025neureg}. Scientific research similarly leverages ontologies as deliberative reasoning layers that anchor neural outputs to verifiable domain knowledge. Across these domains, a common pattern emerges: formal knowledge structures serve as ground truth for neural systems, and retrieval quality directly determines whether such augmentation helps or harms performance.

Mathematical reasoning offers a particularly clean testbed for investigating these dynamics. Competition-level tasks exhibit error rates up to 51.8\% in generated reasoning processes \citep{lightman2023lets}, reflecting broader limitations including hallucination \citep{ji2023survey}, brittleness to perturbations \citep{stolfo2023causal}, and lack of formal grounding \citep{lewkowycz2022solving}. Tool-augmented approaches address these challenges. For instance, MathSensei \citep{das2024mathsensei} achieves 13.5\% improvement through knowledge retrieval and SymPy execution. The most compelling results combine neural pattern recognition with symbolic verification. AlphaGeometry \citep{trinh2024solving} achieves IMO silver medal performance, while rStar-Math \citep{li2025rstar} improves Qwen2.5-Math-7B from 58.8\% to 90.0\% on MATH through code-augmented verification, directly supporting the hypothesis that external grounding enhances reliability.

\subsection{The OpenMath Ontology}
\label{openmath}

OpenMath is a W3C-aligned standard for representing mathematical objects with their semantics, enabling interoperability between mathematical software systems \citep{openmath2019standard}. Unlike presentation formats (e.g., LaTeX), OpenMath focuses on meaning---what expressions represent rather than how they appear. The architecture consists of Content Dictionaries (CDs) serving as modular vocabularies for mathematical domains, with over 200 official and experimental CDs covering arithmetic through advanced algebra, calculus, and logic. Symbol definitions include human-readable descriptions, Commented Mathematical Properties (CMP) in natural language, and Formal Mathematical Properties (FMP) as machine-readable axioms. The Small Type System (STS) provides type signatures enabling arity checking.

For neuro-symbolic AI, OpenMath offers explicit, verifiable formal semantics and mapping to symbolic computation systems (SymPy, Mathematica). However, OpenMath CDs focus primarily on algebraic and analytical mathematics---geometric and combinatorial concepts have limited representation---and significant semantic distance exists between natural language problems and formal definitions, creating retrieval challenges. The strengths and limitations of OpenMath for this application are further analyzed in Sections~\ref{sec:results} and~\ref{sec:discussion}. Complete technical details about the OpenMath Standard are provided in Appendix~\ref{app:appendix_openmath}.

\subsection{Hypothesis on Neuro-Symbolic Ontology-Guided Model Inference}

Based on the literature review and OpenMath's capabilities, I hypothesize that integrating formal mathematical ontologies with language model inference can improve reasoning reliability by reducing hallucination through grounding in verified definitions, enabling verification through mapping to symbolic execution, and improving consistency through anchoring to immutable axioms.

I formalize this as a knowledge-enriched inference framework. Let $\mathcal{P}$ denote a mathematical problem, $\mathcal{K}$ the OpenMath knowledge base, and $\mathcal{M}$ the language model. Standard inference computes:
\begin{equation}
\hat{a} = \mathcal{M}(\mathcal{P})
\end{equation}

An ontology-guided approach modifies this to:
\begin{equation}
\hat{a} = \mathcal{M}(\mathcal{P}, \mathcal{R}(\mathcal{P}, \mathcal{K}))
\end{equation}
where $\mathcal{R}(\mathcal{P}, \mathcal{K})$ retrieves relevant OpenMath definitions for problem $\mathcal{P}$ from knowledge base $\mathcal{K}$. I hypothesize that this augmentation provides benefits when retrieval quality is high, problem types match ontology coverage, and the model has sufficient capacity to process provided definitions. Conversely, performance degradation is expected when irrelevant symbols create noise.

This hypothesis draws support from multiple theoretical perspectives. Dual Process Theory suggests ontologies can guide System 2-like deliberate reasoning that language models (approximating System 1) struggle with \citep{kahneman2011thinking}. Memory-augmented neural networks demonstrate that explicit external memory improves tasks requiring precise recall \citep{graves2016hybrid}. And rStar-Math's success with code execution verification extends naturally to semantic verification through formal ontological properties.

\section{Experiments}
\label{sec:experiments}

\subsection{Project Implementation}

The experiments use the MATH 500, a subset of the MATH benchmark \citep{hendrycks2021measuring} comprising 500 problems across seven mathematical domains and five difficulty levels. Complete benchmark statistics are provided in Appendix~\ref{app:appendix_benchmark}.

The ontology-guided inference system implements a five-phase pipeline illustrated in Figure~\ref{fig:architecture}: knowledge base construction, concept extraction, hybrid retrieval, cross-encoder reranking, and augmented inference. Each phase addresses a specific challenge in bridging natural language problems with formal mathematical definitions. The detailed description of each phase and complete implementation specifications are provided in Appendix~\ref{app:appendix_project_implementation}. The code base and results files are available in this project's Github reposiroty \citep{labre2026openmathpaper}.

\subsection{Model Choice}

As motivated in Section~\ref{sec:introduction}, this work focuses on small language models ($\leq$9B parameters) to investigate whether formal ontological context can help resource-constrained models approach the performance of larger systems. The three models used span the spectrum from minimal capacity to domain expertise. Gemma2-2B (2.6B parameters) represents ``edge AI'' models---efficient but limited in complex reasoning, testing whether small models can leverage external formal knowledge. Gemma2-9B (9.2B parameters) serves as the general-purpose baseline with strong reasoning but no mathematical specialization. And Qwen2.5-Math-7B (7.6B parameters) is a domain expert fine-tuned for mathematical reasoning, investigating the ``expert paradox'' of whether models with strong internal knowledge benefit from---or conflict with---external formal definitions.

\subsection{Experimental Configurations}

Table~\ref{tab:conditions} summarizes the experimental configurations. Detailed prompt templates, threshold coverage statistics, and inference settings are provided in Appendix~\ref{app:experimental_configuration}.

\begin{table}[h]
\centering
\small
\begin{tabular}{@{}p{0.20\textwidth}p{0.75\textwidth}@{}}
\toprule
\textbf{Configuration} & \textbf{Description} \\
\midrule
Conditions & (1) Baseline: models are prompted with only the MATH problem and instructions; (2) OpenMath: Baseline plus retrieved OpenMath context. \\
Threshold Ablation & Model inference for OpenMath reranker thresholds (measure of relevance) varying from 0.0 to 0.9; baseline uses the same filtered set at each threshold. \\
Inference Modes & (1) Greedy: single attempt at temperature zero; (2) Best-of-N: up to 5 attempts at temperature 0.6. \\
\bottomrule
\end{tabular}
\caption{Experimental configurations for comparing baseline and OpenMath.}
\label{tab:conditions}
\end{table}

\subsection{Evaluation Metrics}

Primary evaluation uses answer accuracy---the proportion of problems where the model's extracted answer matches the ground truth: $\text{Accuracy} = \text{Number Correct} / \text{Total Problems}$. To measure improvement from OpenMath augmentation, the percentage accuracy delta computes the difference in percentage points: $\Delta\text{Accuracy} = 100 \times (\text{Accuracy}_{\text{OpenMath}} - \text{Accuracy}_{\text{Baseline}})$, where positive values indicate improvement and negative values indicate degradation.

For best-of-n experiments, $\text{Attempts}$ measures the average number of attempts required to reach a correct answer (or 5 if all fail): $\text{Attempts} = \frac{1}{N}\sum_{i=1}^{N} \text{attempts}_i$. This captures efficiency---whether OpenMath context helps models converge faster. The attempts ratio compares baseline to OpenMath efficiency: $\text{AttemptsRatio} = \text{Attempts}_{\text{Baseline}} / \text{Attempts}_{\text{OpenMath}}$, where ratios $> 1.0$ indicate faster convergence with OpenMath. Finally, the $\Delta\text{Attempts} = \text{Attempts}_{\text{OpenMath}} - \text{Attempts}_{\text{Baseline}}$ indicates the improvement in attempts from OpenMath augmentation.

\section{Results}
\label{sec:results}

This section presents the observations derived from the experimental results. The complete results are available in this project's GitHub repository \citep{labre2026openmathpaper}.

\subsection{Overall Accuracy Impact}

\begin{figure}[!htbp]
\centering
\subfigure[greedy, all levels, all types]{%
    \includegraphics[width=0.45\textwidth]{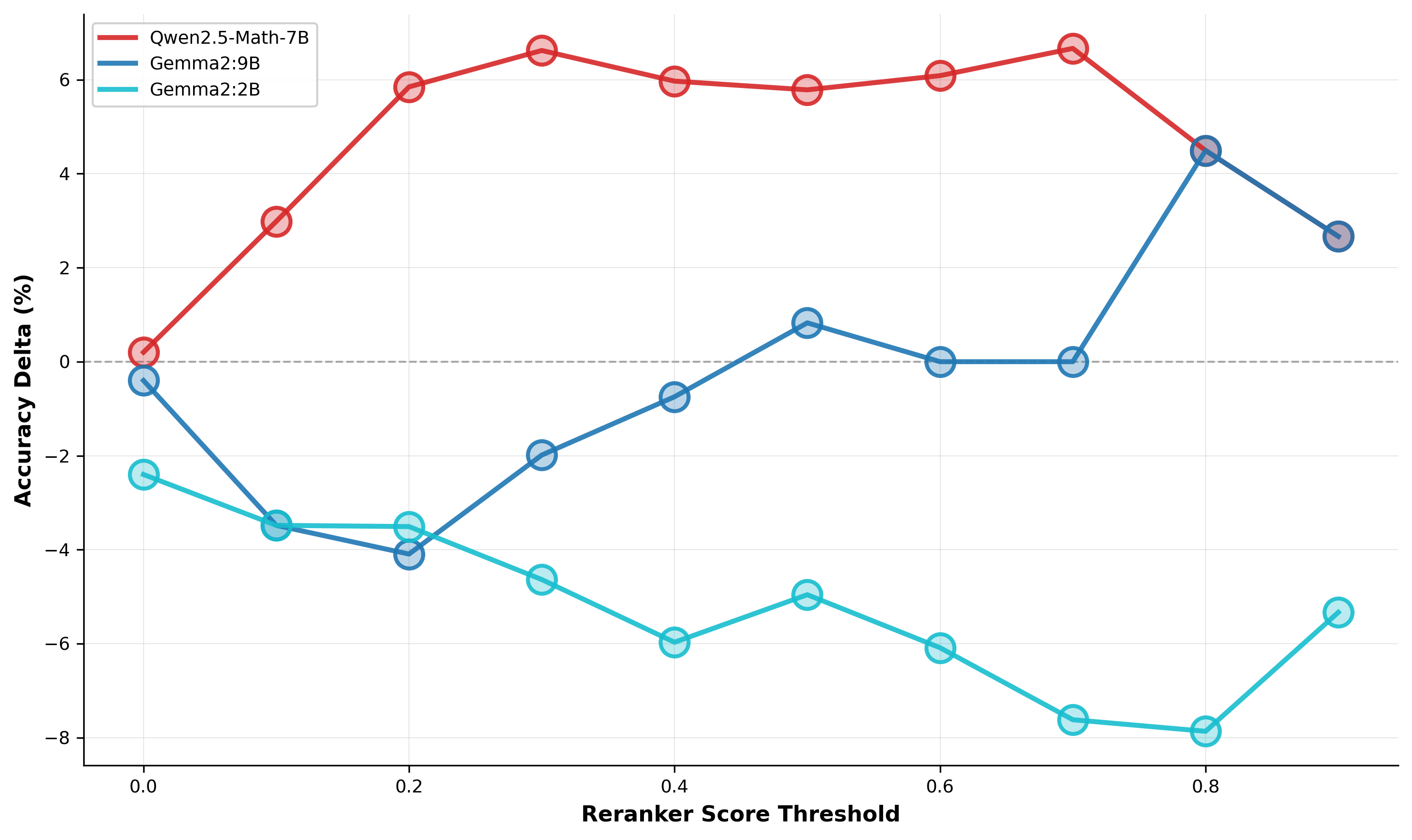}%
}
\subfigure[best-of-n, all levels, all types]{%
    \includegraphics[width=0.45\textwidth]{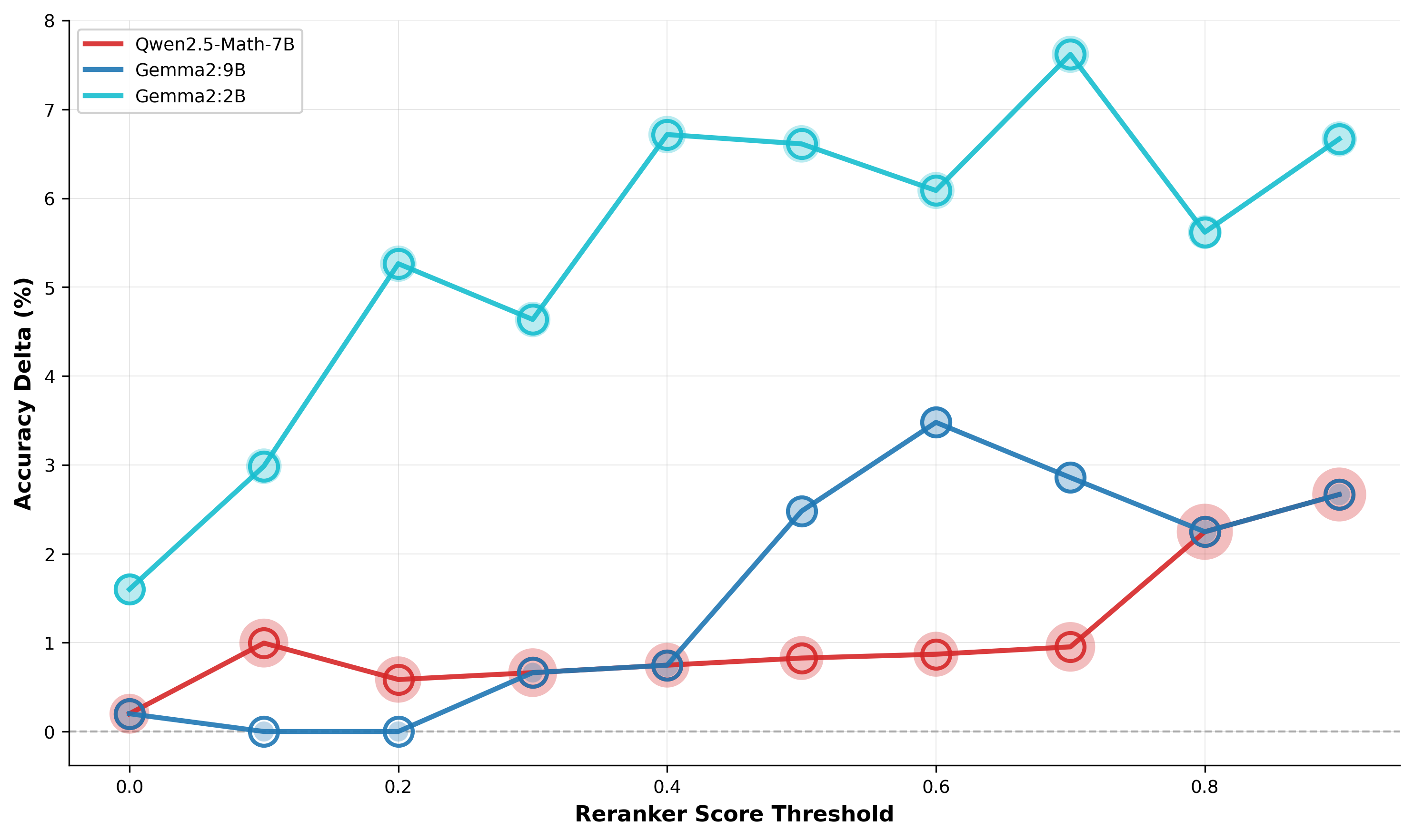}%
}
\caption{$\Delta\text{Accuracy}$ (lines) and $\text{Attempts Ratio}$ (bubbles) by reranker threshold for all MATH 500 problem levels and types.}
\label{fig:overall_accuracy}
\end{figure}

Figure~\ref{fig:overall_accuracy} presents $\Delta\text{Accuracy}$ across reranker score thresholds for all three models, with bubble size measuring $\text{Attempts Ratio}$.

In greedy mode (Figure~\ref{fig:overall_accuracy}a), models exhibit distinct patterns reflecting their architectural differences. Qwen2.5-Math-7B shows consistently positive delta across all thresholds, reflecting its mathematical pre-training that enables effective utilization of formal definitions. Gemma2-9B transitions from slightly negative at low thresholds to positive at higher thresholds, demonstrating that filtering removes noise that initially overwhelms the general-purpose model. Gemma2-2B exhibits consistently negative delta that worsens with higher thresholds, indicating insufficient capacity to utilize formal definitions regardless of filtering stringency.

Best-of-n mode (Figure~\ref{fig:overall_accuracy}b) fundamentally transforms these patterns. All three models achieve positive delta at threshold 0.0. The improvement is most dramatic for Gemma2-2B at threshold 0.7 in best-of-n mode---reversing its greedy-mode degradation. Overall, the convergence of all models toward positive accuracy delta in best-of-n mode suggests that the context recovery mechanism enabled by multiple sampling is broadly beneficial across model architectures and sizes.

\subsection{Accuracy by Difficulty Level}
\label{sec:accuracy_level}

\begin{figure}[!htbp]
\centering
\subfigure[$\Delta\text{Accuracy}$ by level, greedy]{%
    \includegraphics[width=1\textwidth]{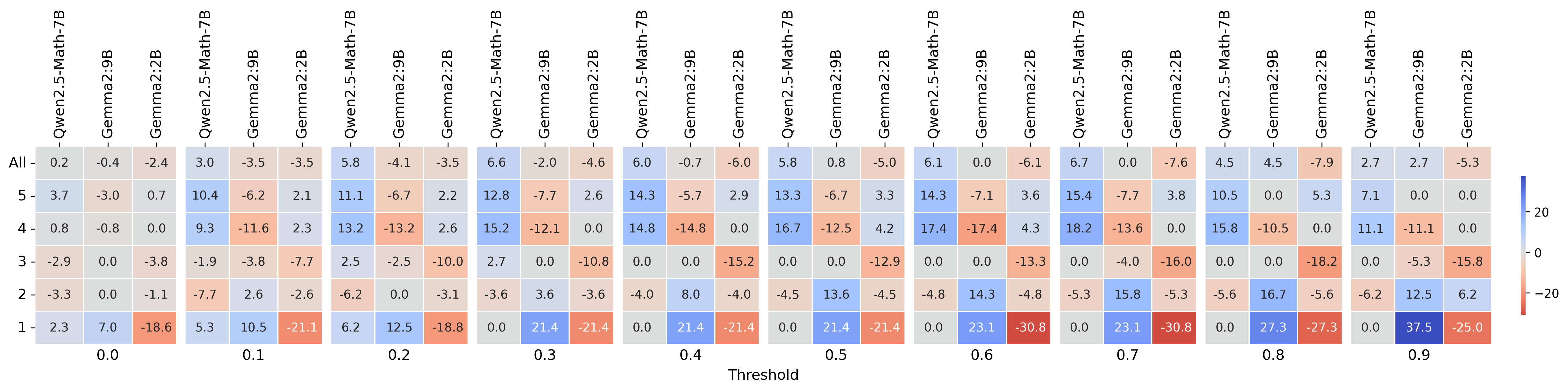}%
}\\
\subfigure[$\Delta\text{Accuracy}$ by level, best-of-n]{%
    \includegraphics[width=1\textwidth]{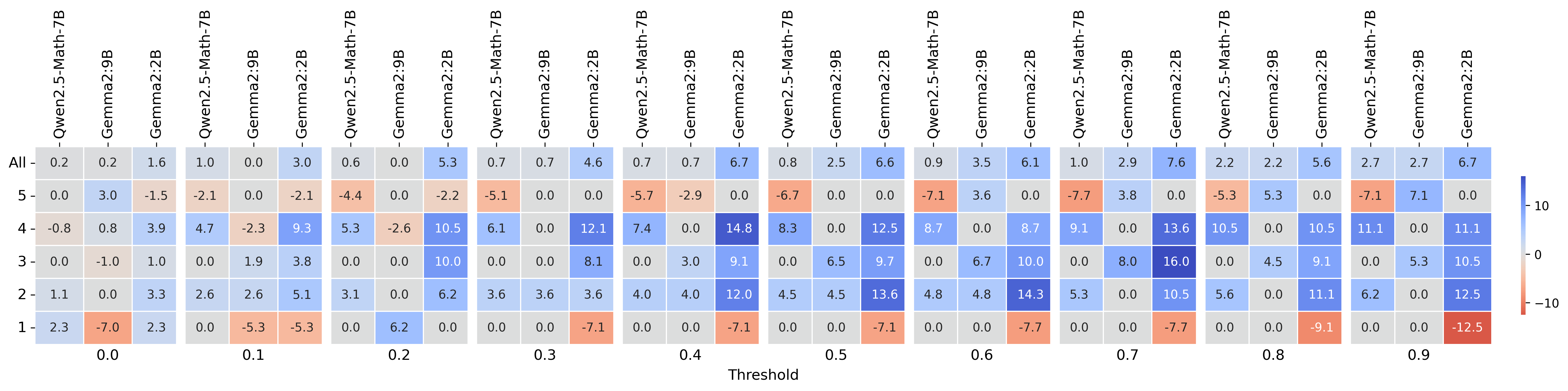}%
}
\caption{$\Delta\text{Accuracy}$ by MATH 500 problem level and reranker threshold. Blue indicates improvement while red indicates degradation.}
\label{fig:accuracy_level}
\end{figure}

Figure~\ref{fig:accuracy_level} presents heatmaps showing $\Delta\text{Accuracy}$ broken down by problem difficulty level (1--5) and reranker score threshold.

Level 1 (easiest) exhibits divergent model behavior at threshold 0.0. Gemma2-9B shows a major improvement in greedy mode, Gemma2-2B degrades significantly, and Qwen2.5-Math-7B achieves a modest gain. The large negative impact on the smallest model reflects context overload on problems where baseline performance is already near-ceiling. The mathematical concepts at this level (basic arithmetic, simple algebra) may already be well-represented in model training data, making additional OpenMath definitions redundant at best and confusing at worst. And yet another contributing factor may be statistical significance (only 43 out 500 problems are level 1: Appendix~\ref{app:appendix_openmath_coverage}).

Levels 3--4 (intermediate to competition-level) show the most consistent benefits for larger models. At Level 4, Gemma2-2B improves substantially in best-of-n mode, while Qwen2.5-Math-7B shows marginal changes at threshold 0.0 but significant improvements at high thresholds. These represent the optimal challenge levels where problems are difficult enough to benefit from external knowledge, yet not so complex that models cannot process the additional context. The formal definitions in OpenMath provide genuine assistance for concepts like GCD, modular arithmetic, and factorials that frequently appear at these difficulty levels.

Level 5 (expert) reveals a \emph{specialization paradox}. At threshold 0.0 in greedy mode, Qwen2.5-Math-7B achieves +3.7\% while Gemma2-9B shows -3.0\%. However, in best-of-n mode, this reverses: Gemma2-9B improves to +3.0\% while Qwen2.5-Math-7B shows 0.0\% at threshold 0.0 and negative values for higher thresholds. The math-specialized model's finely-tuned internal representations may conflict with formal definitions at expert difficulty levels---a parametric-contextual conflict \citep{xu2024knowledge} where learned patterns interfere with external context.

\subsection{Accuracy by Problem Type}

\begin{figure}[!htbp]
\centering
\subfigure[$\Delta\text{Accuracy}$ by type, greedy]{%
    \includegraphics[width=1\textwidth]{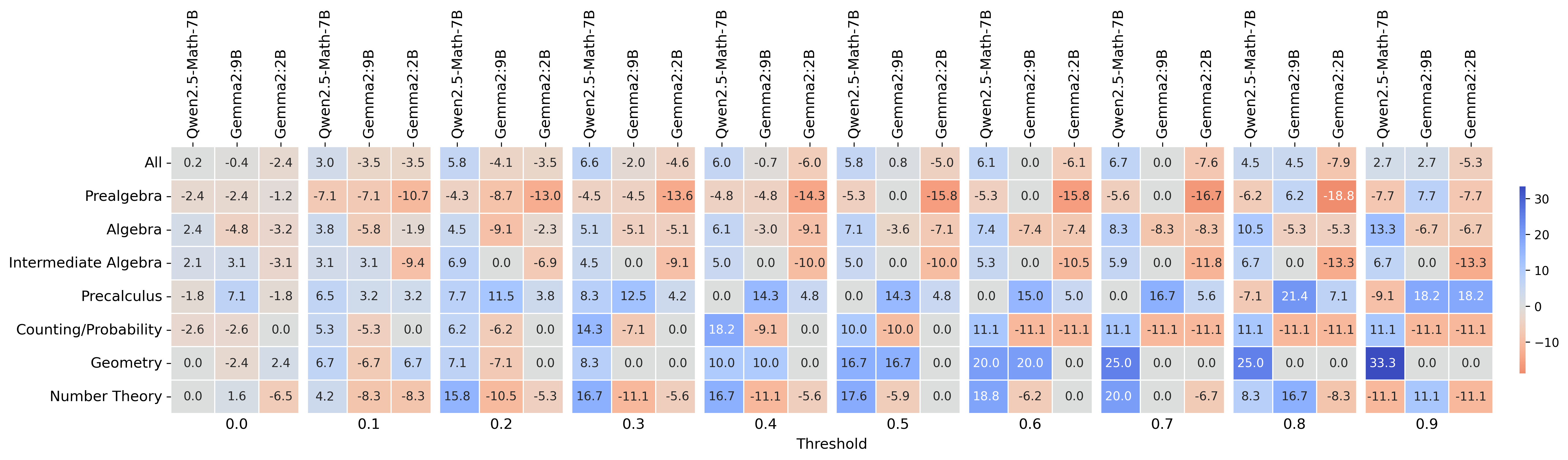}%
}\\
\subfigure[$\Delta\text{Accuracy}$ by type, best-of-n]{%
    \includegraphics[width=1\textwidth]{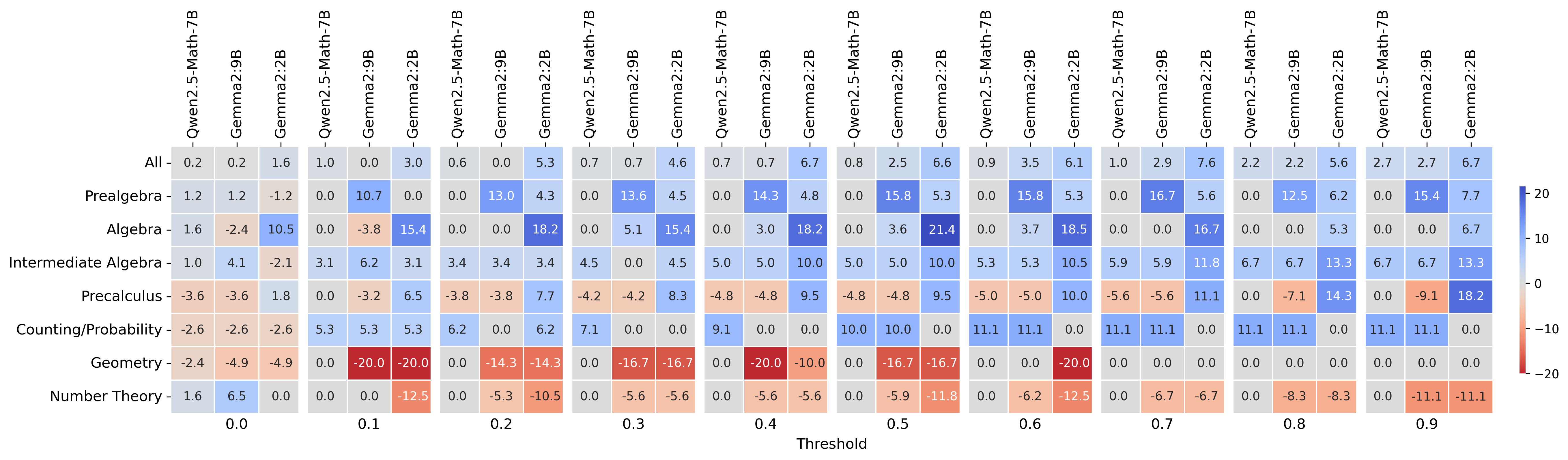}%
}
\caption{$\Delta\text{Accuracy}$ by MATH 500 problem type and reranker threshold. Blue indicates improvement while red indicates degradation.}
\label{fig:accuracy_type}
\end{figure}

Figure~\ref{fig:accuracy_type} presents heatmaps showing $\Delta\text{Accuracy}$ broken down by the seven MATH benchmark problem types.

Algebra demonstrates the strongest positive impact for the math-specialized model. At threshold 0.0 in greedy mode, Qwen2.5-Math-7B achieves +2.4\% on Algebra with improvements increasing to +13.3\% at threshold 0.9. Geometry reaches +33.3\% at threshold 0.9, showing that despite OpenMath's coverage gap in this problem type (Appendix~\ref{app:appendix_openmath_coverage}), this model benefits from the additional higher relevance context.

And the augmentation benefit with Geometry is also apparent in generalist models. Note how both Gemma2:2B and Gemma2:9B recover from poor performance at high thresholds, in both modes. The explanation may be that the highest quality OpenMath coverage in Geometry is concentrated at difficulty levels above 3 (Appendix~\ref{app:appendix_openmath_coverage}). Thus, finer filtering of symbols helps removing noise that otherwise harms model performance in harder problems.

Number Theory has high OpenMath quality coverage, yet both Gemma2 models degrade to -11.1\% at threshold 0.9 in best-of-n mode---suggesting threshold filtering paradoxically harms performance by biasing toward problems where models' internal knowledge suffices. This rationale may be supported by the fact that OpenMath's Number Theory high quality coverage is concentrated in level 1 problems, which collapses from level 2 onward (Appendix~\ref{app:appendix_openmath_coverage}).

Precalculus reveals model-dependent patterns that complicate simple coverage-based explanations. This problem type shows a striking mode-dependent paradox for Gemma2-9B: greedy mode shows improvements increasing with threshold (+7.1\% to +18.2\%), but best-of-n mode reverses this pattern (-3.6\% to -9.1\%), compromising the context recovery hypothesis and suggesting that for some model-domain combinations, multiple sampling introduces counterproductive exploration.

\subsection{Efficiency Analysis: Average Attempts in Best-of-N Mode}
\label{sec:efficiency}

\begin{figure}[!htbp]
\centering
\subfigure[$\Delta\text{Attempts}$ by level, best-of-n]{%
    \includegraphics[width=1\textwidth]{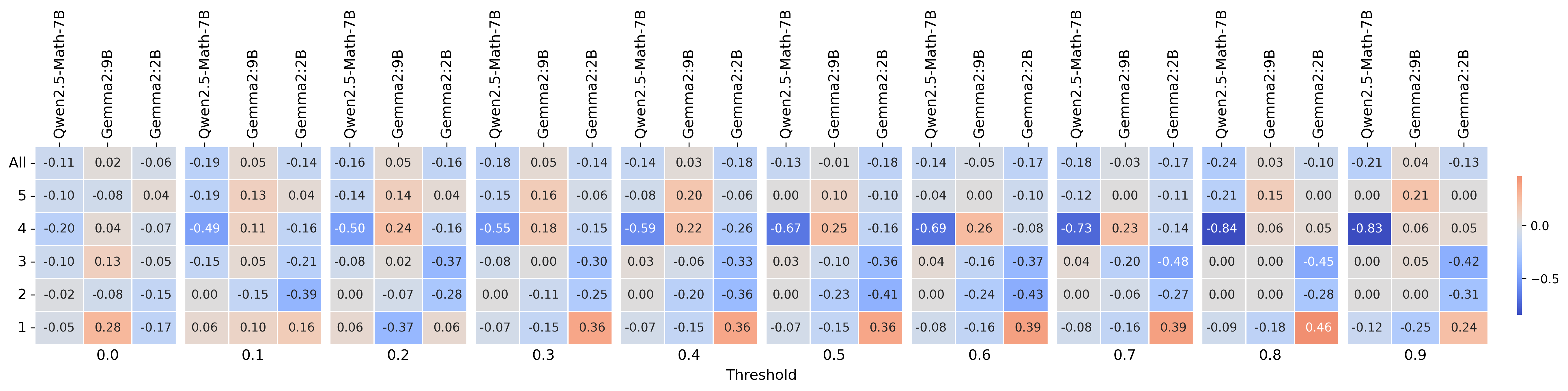}%
}\\
\subfigure[$\Delta\text{Attempts}$ by type, best-of-n]{%
    \includegraphics[width=1\textwidth]{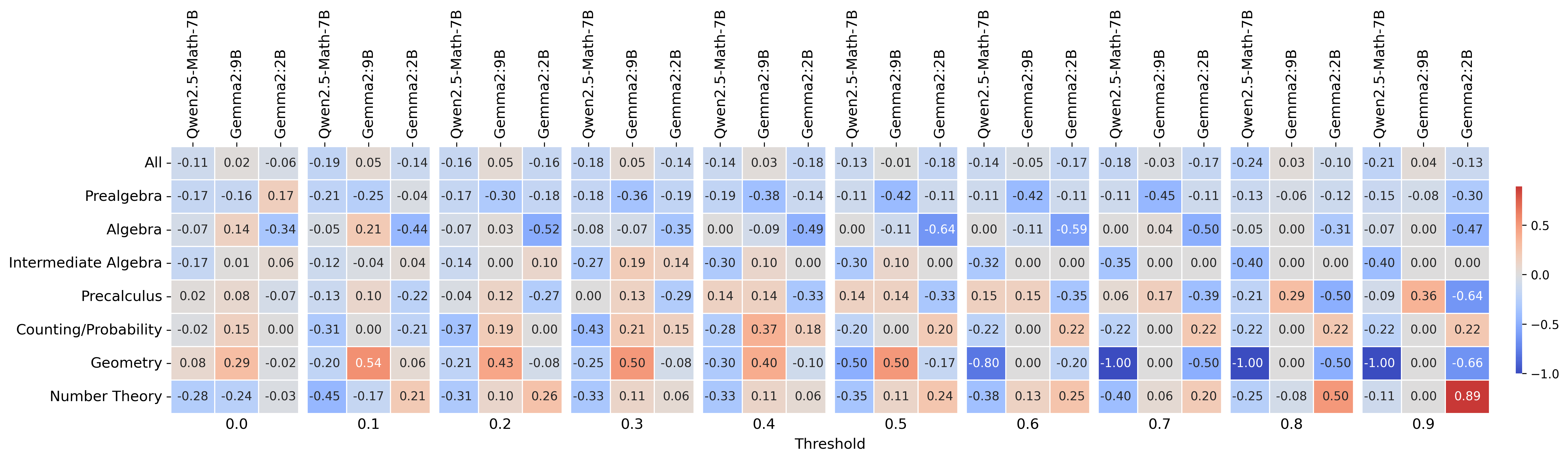}%
}
\caption{$\Delta\text{Attempts}$ by reranker score threshold in best-of-n mode. Negative values (blue) indicate OpenMath requires fewer attempts while positive values (red) indicate more attempts required.}
\label{fig:attempts}
\end{figure}

Figure~\ref{fig:attempts} presents $\Delta\text{Attempts}$ in best-of-n mode where negative values (blue) indicate OpenMath enables fewer attempts, while positive values (red) indicate more attempts required.

In the aggregate of all problem levels and types, OpenMath either helps or has a neutral effect on the number of attempts. This suggests OpenMath provides reasoning shortcuts---formal definitions help models converge to answers somewhat faster, though the effect size is small. The picture becomes more interesting when we break down the results.

The attempts reduction is most pronounced at mid-difficulty levels (Levels 2--4). Optimal improvement at these levels occurs because these problems are challenging enough to benefit from guidance but not so difficult that additional context cannot help. Level 5 problems show smaller reductions because of their higher baseline attempt counts, suggesting that the additional complexity limits the utility of formal definitions for accelerating reasoning.

Note that except for a few statistically less significant cases, OpenMath consistently helps Qwen2.5-Math-7B to solve problems faster compared to baseline. But comparing Figures~\ref{fig:accuracy_level}b and~\ref{fig:attempts}a reveals interesting patterns in terms of efficiency-accuracy dissociation. At Level 5, attempt reduction persists even though accuracy degrades (Section~\ref{sec:accuracy_level}). Reaching answers faster with OpenMath context but without accuracy benefit indicates a potential ``false confidence'' effect, where formal definitions accelerate response generation without improving reasoning quality.

Another interesting pattern emerges with Gemma2-9B on Precalculus, where the model performs more attempts (+0.36 at T=0.9) while achieving worse accuracy (-9.1\%)---a ``false exploration'' effect underscoring that mode selection should perhaps be tailored to specific model-domain combinations. And Number Theory again prompts a model-dependent paradox in generalist models, with Gemma2-2B and Gemma2-9B showing divergent patterns from lower to higher thresholds. These phenomena warrant further investigation.

\section{Discussion}
\label{sec:discussion}

The experimental results from Section~\ref{sec:results} reveal a nuanced picture of ontology-guided inference, where the benefit of formal knowledge augmentation depends critically on the interaction between model characteristics, problem properties, and retrieval quality.

One fundamental finding is the role of model capacity. Gemma2-2B exhibits consistent degradation in greedy mode, Gemma2-9B shows threshold-dependent improvement, and Qwen2.5-Math-7B maintains positive performance throughout. This gradient reflects a ``context utilization ceiling''---smaller models cannot allocate sufficient computational resources to process both problems and formal definitions simultaneously. The dramatic reversal for Gemma2-2B in best-of-n mode reveals that multiple sampling circumvents this ceiling by providing multiple interpretation opportunities.

Threshold analysis demonstrates that retrieval quality is determinative of augmentation success. At low thresholds, generalist models show negative performance due to irrelevant OpenMath symbol noise. At mid-level thresholds, performance improves as obvious noise is filtered while retaining moderately relevant symbols. At very high thresholds, selection bias emerges: problems that draw high relevance OpenMath context are predominantly those where baseline models already perform well, with marginal benefit and decreased statistical power due to smaller sample sizes.

Analysis of the alignment between OpenMath and the MATH 500 benchmark reveals significant coverage gaps: only 24.2\% of problems achieve high-quality coverage (reranker score $\geq$ 0.5), while 65.8\% have poor coverage (score $<$ 0.2). Coverage varies dramatically across domains---Precalculus achieves 37.5\% high-quality coverage, Algebra 22.6\%, Number Theory 27.4\%, while Geometry achieves only 14.6\% (Appendix~\ref{app:appendix_openmath_coverage}). However, the correlation between coverage and performance is not deterministic, as discussed in Section~\ref{sec:results}.

The relationship between problem type performance and OpenMath coverage reveals a few paradoxes. Geometry has the lowest OpenMath coverage, yet all models recover at high thresholds---likely because coverage concentrates at higher difficulty where stricter filtering retains genuinely relevant symbols. Number Theory exhibits the opposite: despite relativelly high coverage, both Gemma2 models degrade at threshold 0.9 in best-of-n mode. Coverage analysis may explain, with high-quality Number Theory coverage concentrated in Level 1 formal problems but collapsing at Level 2 onward where word problems dominate. Precalculus presents a mode-dependent paradox: Gemma2-9B improves in greedy mode but degrades in best-of-n, suggesting context recovery can become counterproductive for certain model-domain combinations.

Divergent patterns at Level 5 reveal a specialization paradox. At threshold 0.0 in greedy mode, Qwen2.5-Math-7B achieves +3.7\% while Gemma2-9B shows -3.0\%. Best-of-n reverses this, with Gemma2-9B improving to +3.0\% and Qwen2.5-Math-7B showing 0.0\% or negative delta. This suggests math-specialized models experience parametric-contextual conflict \citep{xu2024knowledge} at expert levels, where finely-tuned representations interfere with formal definitions. Efficiency analysis supports this: Qwen2.5-Math-7B solves Level 5 problems faster with OpenMath but without accuracy improvement (``false confidence''), while Gemma2-9B on Precalculus requires more attempts while achieving worse accuracy---a ``false exploration'' effect suggesting the need to tailor inference mode to model-domain combinations.

Best-of-n mode generally improves outcomes, supporting a ``context recovery mechanism'' where multiple sampling enables models to overcome initial misinterpretations. This connects to self-consistency research \citep{wang2023selfconsistency}: multiple sampling provides opportunities to re-interpret complex external context. However, this mechanism is not universal: Gemma2-9B on Precalculus shows superior performance in greedy mode compared to best-of-n, as previously observed.

Synthesizing across these themes, OpenMath augmentation can improve mathematical reasoning, but conditionally. Models with fewer than approximately 7B parameters struggle to utilize context in greedy mode. Best-of-n sampling enables smaller models to benefit through context recovery, though not universally. Threshold filtering in the 0.3--0.5 range optimally balances noise reduction and coverage. Performance directly reflects problem-ontology alignment---well-covered types improve while poorly-covered types degrade. Math-specialized models may experience knowledge conflicts at expert levels. And mode selection (greedy vs best-of-n) should be tailored to specific model-domain combinations, as threshold effects can be non-linear and even paradoxical.

The retrieval quality bottleneck emerges as a critical finding: relevant context improves performance while irrelevant context degrades it. Current methods, even with cross-encoder reranking, struggle to bridge the semantic gap between natural language and formal definitions---a limitation affecting all domains where language must map to structured knowledge. For neuro-symbolic AI, these findings offer both validation and caution. Positive results on well-covered types confirm that formal ontologies can ground reasoning when retrieval succeeds. However, negative results demonstrate that augmentation is not universal---it requires knowledge base coverage, retrieval accuracy, model-appropriate injection strategies, and appropriate inference mode selection.

\section{Future Work}

The experimental findings point to a few promising directions for advancing ontology-guided inference.

Improved retrieval methods are an obvious next step in the research, as the retrieval quality bottleneck proved to be the most critical limitation. Semantic retrieval methods that go beyond lexical matching could better capture conceptual relationships between natural language and formal definitions. Hypothetical Document Embeddings (HyDE) \citep{gao2023precise}, where a language model generates idealized matching documents before retrieval, is particularly promising for bridging the semantic gap. Domain-specific rerankers and dynamic threshold selection could further optimize precision.

Beyond prompt-based strategies, mechanistic interpretability offers a promising avenue for aligning models with ontological truth. Techniques such as SAE (Sparse Autoencoder) feature steering \citep{lieberum2024sae} allow precise amplification or suppression of specific internal feature activations during inference. Future work could explore using OpenMath as ground truth to identify and steer feature vectors toward formal mathematical properties.

The framework developed here extends to other specialist domains with established ontologies---medical knowledge graphs, legal ontologies, and scientific domain ontologies---where the lessons learned regarding retrieval quality, coverage alignment, and model capacity transfer directly.

\section{Conclusion}

This work investigated whether formal domain ontologies can enhance language model reasoning reliability through retrieval-augmented generation. Using mathematics as a proof-of-concept domain, I implemented a neuro-symbolic pipeline integrating the OpenMath ontology with multiple language models and evaluated the approach on the MATH benchmark.

The results support a nuanced conclusion: ontology-guided inference can improve performance, but the benefit depends critically on retrieval quality, model capacity, and problem-ontology alignment. When retrieval successfully identifies relevant definitions, accuracy improves; when it fails, performance degrades. The retrieval quality bottleneck emerges as a fundamental challenge to be addressed.

For the broader vision of trustworthy AI in specialist domains, these findings offer both validation and caution---formal knowledge can ground neural reasoning when properly aligned, but ontological augmentation requires careful attention to coverage and model-appropriate injection strategies.

\acks{I would like to express my gratitude for the support of Dell Technologies for providing a Dell Pro Max GB10 workstation to run the experiments of this research. This Pro Max workstation excelled in a demanding experiment pipeline, even with multiple models running concurrently.

During the preparation of this work, the author used Claude and Gemini in order to research the literature, gather information, and improve linguistic clarity and brevity. After using these tools, the author reviewed and edited the content as needed and takes full responsibility for the content of the published article.}

\bibliography{references}

\appendix

\section{The OpenMath Standard}
\label{app:appendix_openmath}

This appendix provides documentation of the OpenMath standard, expanding on the overview in Section~\ref{openmath}. The OpenMath Content Dictionaries (CDs) Github repository (Appendix~\ref{app:appendix_openmath_resources}) was added as a submodule of the infrastructure of this project. Appendix~\ref{app:appendix_openmath_coverage} provides the detailed coverage analysis referenced in Section~\ref{sec:discussion}.

\subsection{Content Dictionary Structure}

Content Dictionaries (CDs) serve as modular vocabularies for specific mathematical domains. Each CD defines a collection of related symbols with their semantics. For example, \texttt{arith1.ocd} defines basic arithmetic operations (\texttt{plus}, \texttt{times}, \texttt{divide}, \texttt{gcd}, \texttt{lcm}), while \texttt{calculus1.ocd} defines differentiation and integration. Symbol Definitions within CDs include:

\begin{itemize}
    \item \textbf{Name}: The symbol identifier (e.g., \texttt{gcd}).
    \item \textbf{Description}: Human-readable explanation.
    \item \textbf{Commented Mathematical Properties (CMP)}: Natural language statements of mathematical properties.
    \item \textbf{Formal Mathematical Properties (FMP)}: Machine-readable axioms and rules expressed in OpenMath XML.
\end{itemize}

\subsection{Example: The GCD Symbol}

Consider the definition of greatest common divisor from \texttt{arith1}:

\begin{verbatim}
Symbol: arith1:gcd
Description: Greatest common divisor of two or more integers
Type: nassoc(SemiGroup) -> SemiGroup
Formal Properties:
  - Commutativity: gcd(a,b) = gcd(b,a)
  - Associativity: gcd(a,gcd(b,c)) = gcd(gcd(a,b),c)
  - Relationship to LCM: gcd(a,b) * lcm(a,b) = a * b
\end{verbatim}

This definition provides not merely a textual description but formal, verifiable properties that govern how GCD behaves mathematically.

\subsection{Small Type System (STS)}

The Small Type System provides lightweight type signatures for mathematical operations, enabling arity and type checking. Types include:

\begin{itemize}
    \item \textbf{Object}: Generic mathematical object.
    \item \textbf{NumericalValue}: Numeric types.
    \item \textbf{mapsto}: Function construction (e.g., \texttt{NumericalValue -> NumericalValue}).
    \item \textbf{nary}: N-ary functions (arbitrary number of arguments).
    \item \textbf{nassoc}: Associative n-ary functions.
\end{itemize}

\subsection{OpenMath and MATH 500 Coverage Analysis}
\label{app:appendix_openmath_coverage}

Understanding the alignment between the OpenMath ontology and the MATH 500 benchmark is essential for interpreting the experimental results. This analysis reveals systematic coverage patterns that directly explain the performance outcomes observed in Section~\ref{sec:results}.

\textbf{Overall Coverage Statistics.} Analysis of all 500 MATH benchmark problems reveals significant coverage gaps in the OpenMath knowledge base. Using the cross-encoder reranker to score semantic relevance between problems and OpenMath symbols, only 24.2\% of problems achieve high-quality coverage (maximum reranker score $\geq$ 0.5), while 65.8\% have poor coverage (score $<$ 0.2). The average maximum relevance score across all problems is 0.2715, indicating that the typical MATH problem finds only marginally relevant matches in OpenMath.

\textbf{Coverage by Problem Type.} The coverage distribution varies dramatically across mathematical domains, as shown in Table~\ref{tab:coverage_type}.

\begin{table}[h]
\centering
\caption{OpenMath coverage by MATH 500 problem type. Coverage is measured by maximum reranker score between problem and best-matching OpenMath symbol.}
\label{tab:coverage_type}
\begin{tabular}{lcccc}
\toprule
Problem Type & Problems & High ($\geq$0.5) & Poor ($<$0.2) & Avg Max Score \\
\midrule
Precalculus & 56 & 37.5\% & 53.6\% & 0.3726 \\
Counting \& Probability & 38 & 26.3\% & 57.9\% & 0.3111 \\
Number Theory & 62 & 27.4\% & 69.4\% & 0.2766 \\
Algebra & 124 & 22.6\% & 64.5\% & 0.2607 \\
Prealgebra & 82 & 23.2\% & 72.0\% & 0.2483 \\
Intermediate Algebra & 97 & 20.6\% & 70.1\% & 0.2321 \\
Geometry & 41 & 14.6\% & 65.9\% & 0.2101 \\
\bottomrule
\end{tabular}
\end{table}

\textbf{Coverage by Problem Level.} Table~\ref{tab:coverage_level} reveals that coverage does not correlate with problem difficulty in an intuitive way. Level 1 (easiest) achieves the best coverage at 32.6\% high-quality, while Level 4 (competition-level) has the worst coverage at only 18.8\% high-quality with 70.3\% poor coverage. This ``coverage valley'' at Level 4 reflects the prevalence of multi-step word problems at this difficulty that do not map well to OpenMath's formal vocabulary.

\begin{table}[h]
\centering
\caption{OpenMath coverage by MATH 500 problem level.}
\label{tab:coverage_level}
\begin{tabular}{lccccc}
\toprule
Level & Problems & High ($\geq$0.5) & Medium (0.2--0.5) & Poor ($<$0.2) & Mean Score \\
\midrule
1 (Easiest) & 43 & 32.6\% & 4.7\% & 62.8\% & 0.3238 \\
2 & 90 & 24.4\% & 11.1\% & 64.4\% & 0.2840 \\
3 & 105 & 29.5\% & 8.6\% & 61.9\% & 0.3091 \\
4 & 128 & 18.8\% & 10.9\% & 70.3\% & 0.2249 \\
5 (Hardest) & 134 & 22.4\% & 11.2\% & 66.4\% & 0.2456 \\
\midrule
\textbf{Total} & 500 & 24.2\% & 10.0\% & 65.8\% & 0.2715 \\
\bottomrule
\end{tabular}
\end{table}

\textbf{Coverage by Level and Type: High-Quality Matrix.} Table~\ref{tab:coverage_matrix_high} presents the percentage of problems achieving high-quality OpenMath coverage ($\geq$0.5) for each Level $\times$ Type combination. The best coverage cells are Level 3 Counting \& Probability (75.0\%, though with small sample size), Level 1 Number Theory (60.0\%), and Level 2 Precalculus (53.8\%). The worst high-coverage cells are Level 1--2 Geometry (0.0\%) and Level 2 Number Theory (10.0\%).

\begin{table}[h]
\centering
\caption{High-quality OpenMath coverage ($\geq$0.5) by problem level and type. Values show percentage (count/total).}
\label{tab:coverage_matrix_high}
\small
\begin{tabular}{lcccccc}
\toprule
Problem Type & L1 & L2 & L3 & L4 & L5 & Row Total \\
\midrule
Algebra & 24\% (4/17) & 19\% (4/21) & 31\% (8/26) & 23\% (7/30) & 17\% (5/30) & 22.6\% \\
Counting \& Prob. & 50\% (1/2) & 43\% (3/7) & 75\% (3/4) & 8\% (1/13) & 17\% (2/12) & 26.3\% \\
Geometry & 0\% (0/2) & 0\% (0/8) & 25\% (2/8) & 20\% (2/10) & 15\% (2/13) & 14.6\% \\
Interm. Algebra & 43\% (3/7) & 17\% (2/12) & 32\% (6/19) & 17\% (4/23) & 14\% (5/36) & 20.6\% \\
Number Theory & 60\% (3/5) & 10\% (1/10) & 12\% (2/16) & 26\% (5/19) & 50\% (6/12) & 27.4\% \\
Prealgebra & 29\% (2/7) & 26\% (5/19) & 29\% (5/17) & 10\% (2/20) & 26\% (5/19) & 23.2\% \\
Precalculus & 33\% (1/3) & 54\% (7/13) & 33\% (5/15) & 23\% (3/13) & 42\% (5/12) & 37.5\% \\
\midrule
\textbf{Column Total} & 32.6\% & 24.4\% & 29.5\% & 18.8\% & 22.4\% & 24.2\% \\
\bottomrule
\end{tabular}
\end{table}

\textbf{Coverage by Level and Type: Poor Coverage Matrix.} Table~\ref{tab:coverage_matrix_poor} presents the percentage of problems with poor OpenMath coverage ($<$0.2). The most problematic cells are Level 2 Number Theory (90.0\% poor), Level 4 Prealgebra (90.0\% poor), and Level 2--3 Geometry (87.5\% poor each).

\begin{table}[h]
\centering
\caption{Poor OpenMath coverage ($<$0.2) by problem level and type. Values show percentage (count/total).}
\label{tab:coverage_matrix_poor}
\small
\begin{tabular}{lcccccc}
\toprule
Problem Type & L1 & L2 & L3 & L4 & L5 & Row Total \\
\midrule
Algebra & 71\% (12/17) & 67\% (14/21) & 58\% (15/26) & 60\% (18/30) & 70\% (21/30) & 64.5\% \\
Counting \& Prob. & 50\% (1/2) & 29\% (2/7) & 25\% (1/4) & 77\% (10/13) & 67\% (8/12) & 57.9\% \\
Geometry & 100\% (2/2) & 88\% (7/8) & 38\% (3/8) & 60\% (6/10) & 69\% (9/13) & 65.9\% \\
Interm. Algebra & 43\% (3/7) & 67\% (8/12) & 63\% (12/19) & 78\% (18/23) & 75\% (27/36) & 70.1\% \\
Number Theory & 40\% (2/5) & 90\% (9/10) & 88\% (14/16) & 68\% (13/19) & 42\% (5/12) & 69.4\% \\
Prealgebra & 71\% (5/7) & 63\% (12/19) & 65\% (11/17) & 90\% (18/20) & 68\% (13/19) & 72.0\% \\
Precalculus & 67\% (2/3) & 46\% (6/13) & 60\% (9/15) & 54\% (7/13) & 50\% (6/12) & 53.6\% \\
\midrule
\textbf{Column Total} & 62.8\% & 64.4\% & 61.9\% & 70.3\% & 66.4\% & 65.8\% \\
\bottomrule
\end{tabular}
\end{table}

\textbf{Geometry.} Geometry achieves the lowest coverage of any problem type (14.6\% high-quality, 65.9\% poor), with particularly severe gaps at basic difficulty levels. Level 1 and Level 2 Geometry problems achieve 0\% high-quality coverage, with 100\% and 87.5\% poor coverage respectively. Only at Level 3 does any meaningful coverage emerge (25\% high-quality). This pattern reflects fundamental gaps in OpenMath's Content Dictionaries. The geometry CDs (\texttt{plangeo1-5}) focus on incidence geometry---relationships between points, lines, and their intersections---rather than the Euclidean metric geometry tested in MATH 500. Key missing concepts include standard shapes (polygon, hexagon, triangle, quadrilateral), perimeter and circumference calculations, and area formulas for common figures. The experimental results in Section~\ref{sec:results} show that Geometry consistently underperforms across all models, with generalist models experiencing substantial degradation---a direct consequence of this coverage gap where retrieved symbols introduce noise rather than guidance.

\textbf{Precalculus.} Precalculus achieves the highest coverage of any problem type (37.5\% high-quality, 53.6\% poor), benefiting from OpenMath's strong representation of trigonometric functions in the \texttt{transc1} Content Dictionary. Level 2 Precalculus achieves 53.8\% high-quality coverage---the best single-cell coverage in the entire matrix---explaining why this level-type combination shows positive accuracy delta for most models. The coverage pattern for Precalculus is notably consistent across difficulty levels (33\%--54\% high-quality), unlike other problem types that show dramatic level-dependent variation.

\textbf{Number Theory.} Number Theory exhibits the most striking level-dependent coverage pattern. Level 1 Number Theory problems achieve 60.0\% high-quality coverage with only 40.0\% poor coverage, while Level 2 collapses to just 10.0\% high-quality coverage with 90.0\% poor coverage---an 83\% relative decline. Level 3 continues this pattern with only 12.5\% high-quality coverage. This collapse occurs because Level 1 Number Theory problems are stated formally (e.g., ``What is gcd(48, 18)?'' or ``What is 2004 mod 12?''), mapping directly to OpenMath's \texttt{arith1} Content Dictionary symbols. In contrast, Levels 2--3 are dominated by word problems involving ages, earnings, counting, and relationships that require extracting mathematical operations from natural language context. OpenMath's formal vocabulary provides no representation for word problem primitives (rate, proportion, percent, age relationships). Interestingly, Level 5 Number Theory recovers to 50.0\% high-quality coverage because expert-level problems return to formal mathematical statements involving proofs, modular systems, and advanced number-theoretic concepts that align with OpenMath's formal definitions.

\textbf{Structural Limitations.} These coverage patterns reflect that OpenMath was designed as an interchange format for computer algebra systems, not as a knowledge base for natural language reasoning \citep{kohlhase2012semantics}. Effective ontology-guided inference for mathematical reasoning would require either expanding OpenMath's vocabulary to include word problem primitives and Euclidean geometry concepts, or developing adaptive injection strategies that detect problem characteristics and skip augmentation when coverage is guaranteed to fail.

\subsection{References and Resources}
\label{app:appendix_openmath_resources}

\begin{itemize}
    \item \textbf{The OpenMath Standard}: \url{https://openmath.org/standard/om20-2019-07-01/omstd20.html}
    \item \textbf{Small Type System}: \url{https://openmath.org/standard/sts.pdf}
    \item \textbf{Content Dictionaries}: \url{https://github.com/OpenMath/CDs}
\end{itemize}

\section{Project Implementation}
\label{app:appendix_project_implementation}

This appendix provides dataset statistics and complete implementation details for the ontology-guided inference pipeline.

\subsection{Benchmark}
\label{app:appendix_benchmark}

I evaluate the approach on the MATH benchmark \citep{hendrycks2021measuring}, using the standard MATH 500 subset employed in recent literature \citep{lightman2023lets,li2025rstar,shao2024deepseekmath}. Tables~\ref{tab:domain_dist} and~\ref{tab:level_dist} summarize the distribution across seven mathematical domains and five difficulty levels, with harder problems (levels 4--5) comprising 52.4\% of the dataset.

\begin{table}[h]
\centering
\begin{minipage}{0.48\textwidth}
\centering
\caption{Problems by domain.}
\label{tab:domain_dist}
\begin{tabular}{lc}
\toprule
Type & N (\%) \\
\midrule
Algebra & 124 (24.8\%) \\
Intermediate Algebra & 97 (19.4\%) \\
Number Theory & 62 (12.4\%) \\
Prealgebra & 82 (16.4\%) \\
Precalculus & 56 (11.2\%) \\
Geometry & 41 (8.2\%) \\
Counting \& Prob. & 38 (7.6\%) \\
\bottomrule
\end{tabular}
\end{minipage}
\hfill
\begin{minipage}{0.48\textwidth}
\centering
\caption{Problems by level.}
\label{tab:level_dist}
\begin{tabular}{lc}
\toprule
Level & N (\%) \\
\midrule
1 (Basic) & 43 (8.6\%) \\
2 (Intermediate) & 90 (18.0\%) \\
3 (Advanced) & 105 (21.0\%) \\
4 (Competition) & 128 (25.6\%) \\
5 (Expert) & 134 (26.8\%) \\
\bottomrule
\end{tabular}
\end{minipage}
\end{table}

Each problem includes a ground truth answer in \verb|\boxed{}| notation, which may be numeric (integers, fractions, decimals), symbolic (algebraic expressions, equations), or compound (tuples, sets). Answer verification employs exact string matching after normalization, with SymPy symbolic equivalence checking for algebraic expressions.

\subsection{Pipeline Overview}

The ontology-guided inference system implements a multi-stage pipeline transforming mathematical problems into augmented prompts:

\begin{itemize}
    \item \textbf{Phase 1: Knowledge Base Construction.} Parse OpenMath CDs into a structured JSON knowledge base with normalized definitions and type signatures.
    \item \textbf{Phase 2: Concept Extraction.} Use a language model to extract mathematical concepts from natural language problems.
    \item \textbf{Phase 3: Hybrid Retrieval.} Combine BM25 (lexical) and dense embedding (semantic) retrieval with Reciprocal Rank Fusion (RRF) to retrieve candidate symbols.
    \item \textbf{Phase 4: Cross-Encoder Reranking.} Apply a reranker model to filter irrelevant candidates and score relevance.
    \item \textbf{Phase 5: Augmented Inference.} Format relevant OpenMath definitions for prompt injection, construct prompt and generate responses with ontological context.
\end{itemize}

This architecture enables systematic investigation of each component's contribution to overall performance.

\begin{figure}[!htbp]
\centering
\includegraphics[width=\textwidth]{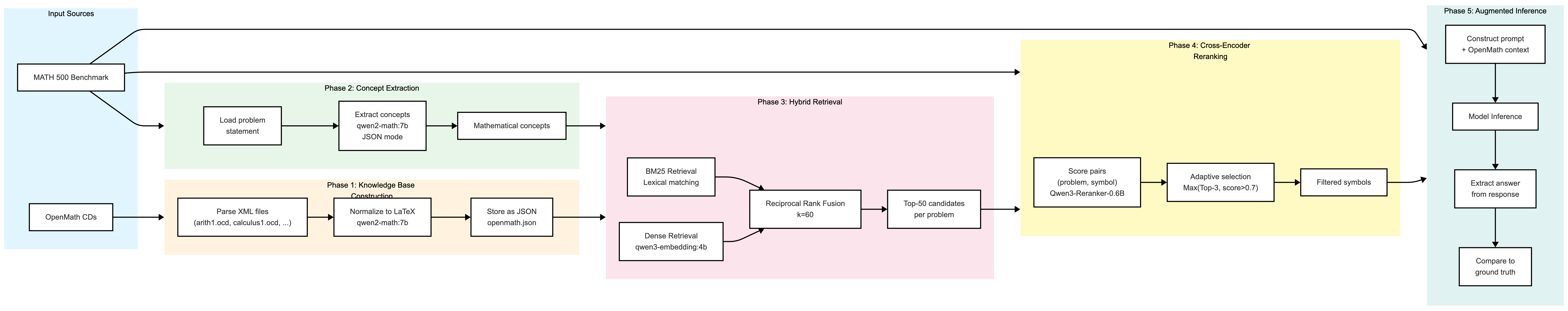}
\caption{System architecture for ontology-guided mathematical inference.}
\label{fig:architecture}
\end{figure}

\subsection{Knowledge Base Construction}

The OpenMath Content Dictionaries store mathematical expressions in a variety of formats---some in plain text (e.g., \texttt{gcd(6,9) = 3}), others using lambda notation (\texttt{diff(lambda y:integral(lambda z:f(z))(y)) = f}), and many with mixed natural language and mathematical expressions. This heterogeneity poses challenges for both retrieval (inconsistent indexing) and prompt construction (cognitive load on the language model). To address this, I implemented a \textbf{normalization pipeline} that converts all mathematical expressions to consistent LaTeX format.

\textbf{Symbol entries.} The knowledge base construction process parses 156 Content Dictionaries containing 1,138 mathematical symbols. Each symbol entry includes:

\begin{itemize}
    \item \textbf{Symbol identifier.} Unique \texttt{cd:name} reference (e.g., \texttt{arith1:gcd})
    \item \textbf{Description.} Human-readable explanation of the symbol's semantics.
    \item \textbf{Properties.} Commented Mathematical Properties (CMP)---natural language statements of mathematical properties.
    \item \textbf{Type signature.} From the Small Type System (STS) specifying arity and types.
    \item \textbf{SymPy mapping.} Where available, the corresponding SymPy function for execution.
\end{itemize}

\textbf{Normalization Pipeline.} The normalization employs a hybrid approach with two stages:

\begin{itemize}
    \item \textbf{Pattern-based conversion} (fast path). Regex patterns detect common mathematical expressions---lambda expressions, function applications, arithmetic operators---and convert them directly to LaTeX using SymPy's \texttt{latex()} function.
    \item \textbf{Model fallback} (slow path). Expressions that fail pattern-based conversion are processed by \texttt{qwen2-math:7b} via Ollama. The model receives few-shot examples demonstrating correct LaTeX conversion and produces structured output. A validation step ensures the output contains valid LaTeX before acceptance.
\end{itemize}

Example conversions:

\begin{table}[h]
\centering
\caption{Example normalization conversions from raw OpenMath format to LaTeX.}
\label{tab:normalization}
\begin{tabular}{p{6cm}p{6cm}}
\toprule
Original & Normalized \\
\midrule
\texttt{diff(lambda y:integral(lambda z:f(z))(y)) = f} & $\frac{d}{dy}(\int f(z) \, dz) = f$ \\
\texttt{sin(x)\^{}2 + cos(x)\^{}2 = 1} & $\sin^{2}(x) + \cos^{2}(x) = 1$ \\
\texttt{If is\_commutative(G) then for all a,b in carrier(G) a*b = b*a} & $\text{is\_commutative}(G) \Rightarrow \forall a, b \in \text{carrier}(G): a \cdot b = b \cdot a$ \\
\bottomrule
\end{tabular}
\end{table}

\textbf{Manual Corrections.} Automated normalization achieved 82\% coverage, but 195 entries required manual correction due to: SymPy parsing failures producing garbled output (tokenizing function names letter-by-letter), right-hand side truncation in equations, unit symbol garbling (e.g., ``minute'' parsed as product of single-letter variables), and mid-expression quantifiers (``for all'') not converted to $\forall$.

These corrections were documented and applied via a reproducible script, ensuring the knowledge base can be regenerated if needed.

\textbf{Final Statistics.} The pipeline filters 98 non-mathematical symbols (metadata, protocol definitions), yielding 1,040 indexed symbols with normalized descriptions, properties, and examples stored in \texttt{*\_normalized} JSON fields alongside the originals.

\subsection{Concept Extraction}

A fundamental challenge in ontology-guided inference is bridging the semantic gap between natural language problem statements and formal ontological definitions. MATH benchmark problems are expressed in verbose natural language with contextual details, while OpenMath symbols are defined using precise mathematical terminology. Directly matching problem text against symbol descriptions yields poor retrieval quality.

To address this, I implemented a concept extraction stage that transforms each problem into a clean list of mathematical keywords and operators. The extracted concepts serve as an intermediate representation optimized for matching against OpenMath symbol definitions in the subsequent retrieval stage.

\textbf{Extraction Approach.} I use \texttt{qwen2-math:7b}, a mathematics-specialized language model, served via Ollama in structured JSON mode. The model is instructed to extract mathematical concepts \emph{without solving the problem}:

\begin{verbatim}
You are a mathematical entity extractor. Extract the core mathematical
concepts from a problem WITHOUT solving it.

Extract these types of concepts:
- Operations: addition, integration, differentiation
- Functions: gcd, sin, log, factorial, determinant
- Objects: integer, polynomial, matrix, set, sequence
- Domains: algebra, calculus, number theory, combinatorics

Return ONLY a JSON object: {"concepts": ["...", "..."]}
\end{verbatim}

The mathematics-specialized model outperforms general-purpose models for this task because it recognizes mathematical terminology and operations that general models might miss or misclassify. A token limit (100 tokens) prevents the model from attempting to solve the problem.

\textbf{Statistics.} Across the 500-problem test set:
\begin{itemize}
    \item Total concepts extracted: 3,004.
    \item Average concepts per problem: 6.0.
    \item Problems in target range (4--8 concepts): 453 (91\%).
    \item Manual corrections required: 1 problem (0.2\%).
\end{itemize}

\subsection{Hybrid Retrieval}

Given extracted concepts from a problem and 1,040 indexed OpenMath symbols, the retrieval stage identifies the most relevant symbol candidates. I implement a hybrid retrieval approach that combines lexical and semantic methods, addressing complementary failure modes: lexical methods miss semantically related terms (e.g., ``GCD'' vs ``greatest common divisor''), while semantic methods may return contextually similar but mathematically irrelevant symbols.

\textbf{BM25 (Lexical Retrieval).} The BM25 algorithm \citep{robertson1995okapi} provides term-frequency-based matching between extracted concepts and symbol ``description cards''---concatenated text comprising the symbol's description, normalized properties, and examples. BM25 excels at exact term matches: when a problem mentions ``factorial,'' BM25 ranks \texttt{integer1:factorial} highly due to direct term overlap.

Key preprocessing steps: Tokenization with mathematical stopword removal (70+ stopwords including ``the,'' ``is,'' ``function,'' common mathematical verbs). The description card uses normalized fields (\texttt{description\_normalized}, \texttt{cmp\_properties\_normalized}, \texttt{examples\_normalized}) to ensure consistent LaTeX formatting.

\textbf{Dense Embedding (Semantic Retrieval).} For semantic similarity, I generate dense embeddings using \texttt{qwen3-embedding:4b} via the Ollama API. Each OpenMath symbol's description card is embedded into a 2,560-dimensional vector space. At query time, the concatenated concepts are embedded and compared via cosine similarity.

Dense retrieval captures semantic relationships beyond lexical overlap:
\begin{itemize}
    \item ``Derivative'' matches \texttt{calculus1:diff} even without the exact term.
    \item ``Roots of polynomial'' retrieves \texttt{polynomial4:factors} through learned semantic associations.
\end{itemize}

\textbf{Embedding Cache.} To enable efficient batch processing, embeddings are pre-computed and cached:
\begin{itemize}
    \item OpenMath symbol embeddings:\newline
    \path{data/openmath-embeddings_qwen3-embedding_4b.npy} (1,040 $\times$ 2,560).
    \item MATH 500 concept embeddings:\newline
    \path{data/math500-concepts-embeddings_qwen3-embedding_4b.npy} (500 $\times$ 2,560).
\end{itemize}

\textbf{Reciprocal Rank Fusion (RRF).} Results from both methods are combined using Reciprocal Rank Fusion \citep{cormack2009reciprocal}, a robust rank aggregation technique:
\begin{equation}
\text{RRF}(d) = \sum_{r \in R} \frac{w_r}{k + \text{rank}_r(d)}
\end{equation}
where $R$ is the set of ranking methods, $w_r$ is the weight for method $r$, $k$ is a smoothing constant (set to 60), and $\text{rank}_r(d)$ is the rank of document $d$ in method $r$'s results. I use equal weights ($w_{BM25} = w_{Dense} = 0.5$) for the fusion. RRF is preferred over score-based fusion because it is robust to score distribution differences between methods.

\textbf{Output.} The top 50 candidates by RRF score are retained for each problem, yielding 25,000 total candidate (problem, symbol) pairs across the 500-problem dataset. This high recall setting (50 candidates) ensures relevant symbols are not missed, delegating precision improvement to the cross-encoder reranking stage.

\subsection{Cross-Encoder Reranking}

While hybrid retrieval achieves high recall, many of the 50 candidates per problem are irrelevant---a phenomenon I term ``domain overreach,'' where symbols from adjacent mathematical domains are retrieved based on superficial similarity. For example, a problem mentioning ``greatest common divisor'' might retrieve both \texttt{arith1:gcd} (integer GCD, highly relevant) and \texttt{polynomial3:gcd} (polynomial GCD, less relevant for integer problems). A cross-encoder reranker addresses this by scoring each (problem, symbol) pair for semantic relevance.

\textbf{Reranker Model.} I employ Qwen3-Reranker-0.6B \citep{qwen2025reranker}, a 600M parameter cross-encoder model specifically designed for relevance scoring. The model is served via vLLM's pooling endpoint, which provides native cross-encoder support through the \texttt{/score} API. This architecture---rather than prompting a language model for relevance judgments---enables efficient batch scoring of 25,000 pairs.

The vLLM server requires specific configuration to enable proper sequence classification:
\begin{verbatim}
vllm serve Qwen/Qwen3-Reranker-0.6B --runner pooling --port 9001 \
  --hf-overrides '{"architectures":["Qwen3ForSequenceClassification"],
                   "classifier_from_token":["no","yes"],
                   "is_original_qwen3_reranker":true}'
\end{verbatim}

\textbf{Scoring Process.} For each (problem, symbol) pair:
\begin{itemize}
    \item \textbf{Input.} Problem statement concatenated with symbol description card.
    \item \textbf{Output.} Relevance score in range [0.0, 1.0].
    \item \textbf{High Scores.} ($>$0.8) indicate strong semantic alignment.
    \item \textbf{Low scores.} ($<$0.2) indicate domain overreach or irrelevance.
\end{itemize}

\textbf{Symbol Selection Rule.} Rather than applying a fixed threshold, I use an adaptive selection rule that balances coverage and precision:
\begin{equation}
\text{Selected} = \max\left(\text{Top-}3_{\text{RRF}}, \; \{s : \text{reranker\_score}(s) > 0.7\}\right)
\end{equation}

This rule ensures:
\begin{itemize}
    \item \textbf{Minimum coverage.} At least 3 symbols are retained (the top-3 by RRF score), ensuring problems with weak OpenMath coverage still receive some context.
    \item \textbf{Quality preference.} If more than 3 symbols score above 0.7, all high-scoring symbols are retained.
    \item \textbf{Graceful degradation.} Problems with poor OpenMath matches still proceed to inference, enabling threshold-based analysis.
\end{itemize}

\textbf{Statistics:}
\begin{itemize}
    \item Total (problem, symbol) pairs scored: 25,000.
    \item Pairs retained after filtering: 1,543 (6.2\% retention).
    \item Average symbols retained per problem: 3.1.
    \item High-confidence matches (score $>$ 0.8): 89 problems (18\%).
    \item Low-confidence matches (max score $<$ 0.2): 261 problems (52\%).
\end{itemize}

The high proportion of low-confidence matches (52\%) reflects systematic coverage gaps in OpenMath for geometry and combinatorics problems, a limitation explored in Appendix~\ref{app:appendix_openmath_coverage}.

\subsection{Implementation Details}

The experiments were conducted on a Dell Pro Max workstation with NVIDIA Blackwell GB10 GPU. The key implementation details are:

\begin{itemize}
    \item \textbf{Inference Framework.} Ollama API.
    \item \textbf{Embedding Model.} qwen3-embedding:4b via Ollama API.
    \item \textbf{Reranker.} Qwen3-Reranker-0.6B via vLLM pooling endpoint.
    \item \textbf{Answer Extraction.} Regex-based extraction of \verb|\boxed{}| content with fallback patterns.
    \item \textbf{Answer Comparison.} Exact match after normalization, with SymPy symbolic equivalence for expressions.
\end{itemize}

\section{Experimental Configuration}
\label{app:experimental_configuration}

This appendix provides detailed experimental configurations referenced in Section~\ref{sec:experiments}.

\subsection{Prompt Templates}

\textbf{Baseline Condition.} The model receives only the problem statement:

\begin{verbatim}
Problem: [MATH problem statement]

Solve this problem step by step. Put your final answer in \boxed{}.
\end{verbatim}

\textbf{OpenMath Condition.} The model receives augmented context:

\begin{verbatim}
The following mathematical definitions may be relevant:

Symbol: arith1:gcd
Description: The greatest common divisor of two or more integers.
Properties: gcd(a,b) = gcd(b,a); gcd(a,gcd(b,c)) = gcd(gcd(a,b),c)

[Additional symbols as retrieved...]

Problem: [MATH problem statement]

Use the provided definitions if relevant. Solve step by step.
Put your final answer in \boxed{}.
\end{verbatim}

The prompt explicitly allows the model to ignore definitions if irrelevant (``Use the provided definitions \emph{if relevant}''), providing an escape hatch for cases where retrieval returns poor matches.

\subsection{Threshold Coverage and Inference Configuration}

\begin{table}[h]
\centering
\begin{minipage}[t]{0.48\textwidth}
\centering
\caption{Threshold levels and corresponding problem coverage.}
\begin{tabular}{lcc}
\toprule
Threshold & Problems & Coverage \\
\midrule
0.0 & 500 & 100\% \\
0.1 & 201 & 40.2\% \\
0.3 & 151 & 30.2\% \\
0.5 & 121 & 24.2\% \\
0.7 & 105 & 21.0\% \\
0.9 & 75 & 15.0\% \\
\bottomrule
\end{tabular}
\end{minipage}
\hfill
\begin{minipage}[t]{0.48\textwidth}
\centering
\caption{Inference parameters by mode.}
\begin{tabular}{lcc}
\toprule
Parameter & Greedy & Best-of-N \\
\midrule
Temperature & 0.0 & 0.6 \\
Max attempts & 1 & 5 \\
\bottomrule
\end{tabular}
\end{minipage}
\end{table}

\end{document}